%% file: main.tex
\documentclass[letterpaper]{article} 
\usepackage{aaai23}  
\usepackage{times}  
\usepackage{helvet}  
\usepackage{courier}  
\usepackage[hyphens]{url}  
\usepackage{graphicx} 
\usepackage{natbib}  
\usepackage{caption} 
\usepackage{algorithm}
\usepackage{algorithmic}

\usepackage{url}            
\usepackage{booktabs}       
\usepackage{amsfonts}       
\usepackage{nicefrac}       
\usepackage{microtype}      
\usepackage{xcolor}         
\usepackage{subcaption}
\usepackage{siunitx}
\usepackage{amsthm}
\usepackage{amsmath}
\usepackage{bbold}
\usepackage[hang,flushmargin]{footmisc}

\usepackage{newfloat}
\usepackage{listings}
\DeclareCaptionStyle{ruled}{labelfont=normalfont,labelsep=colon,strut=off} 
\floatstyle{ruled}
\newfloat{listing}{tb}{lst}{}
\floatname{listing}{Listing}
\newtheorem{theorem}{Theorem}
\newtheorem{corollary}{Corollary}
\DeclareMathOperator*{\argmax}{argmax}

\newcommand\blfootnote[1]{%
  \begingroup
  \renewcommand\thefootnote{}\footnote{#1}%
  \addtocounter{footnote}{-1}%
  \endgroup
}

\title{Learning the Finer Things: Bayesian Structure Learning at the Instantiation Level}
\author{
    Chase Yakaboski,
    Eugene Santos, Jr.
}
\affiliations{
    Thayer School of Engineering\\
    Dartmouth College\\
    Hanover, NH 03755 \\
    \texttt{chase.th@dartmouth.edu, esj@dartmouth.edu} \\
}

\begin{document}
\maketitle

\begin{abstract}
Successful machine learning methods require a trade-off between memorization and generalization. Too much memorization and the model cannot generalize to unobserved examples. Too much over-generalization and we risk under-fitting the data. While we commonly measure their performance through cross validation and accuracy metrics, how should these algorithms cope in domains that are extremely under-determined where accuracy is always unsatisfactory? We present a novel probabilistic graphical model structure learning approach that can learn, generalize and explain in these elusive domains by operating at the random variable instantiation level. Using Minimum Description Length (MDL) analysis, we propose a new decomposition of the learning problem over all training exemplars, fusing together minimal entropy inferences to construct a final knowledge base. By leveraging Bayesian Knowledge Bases (BKBs), a framework that operates at the instantiation level and inherently subsumes Bayesian Networks (BNs), we develop both a theoretical MDL score and associated structure learning algorithm that demonstrates significant improvements over learned BNs on 40 benchmark datasets. Further, our algorithm incorporates recent off-the-shelf DAG learning techniques enabling tractable results even on large problems. We then demonstrate the utility of our approach in a significantly under-determined domain by learning gene regulatory networks on breast cancer gene mutational data available from The Cancer Genome Atlas (TCGA).
\end{abstract}

\section{Introduction} \label{introduction}
\blfootnote{Github URL: \url{https://github.com/di2ag/pybkb}}
Since popularization by Pearl \shortcite{pearl_fusion_1986}, learning Bayesian Networks (BNs) has solidified into a steadfast research area for 40 years. It has become an important paradigm for modeling and reasoning under uncertainty and has seen applications from stock market prediction \cite{malagrino_forecasting_2018} and medical diagnosis \cite{shih_formal_2018} to Gene Regulatory Networks (GRNs) \cite{sauta_bayesian_2020}. Despite Bayesian Network Structure Learning (BNSL) being NP-hard \cite{chickering_large-sample_2004} and even simpler structures like polytrees being NP-hard(er) \cite{dasgupta_learning_1999}, new constraints \cite{gruttemeier_learning_2020}, improvements \cite{trosser_improved_2021}, and scalings \cite{scanagatta_learning_2015} are presented at major AI conferences every year. This is because BNs and affliated structures like Markov \cite{koller_probabilistic_2009} and Dependency Networks (DNs) \cite{heckerman_dependency_2000} offer a quality that other methods such as deep learning do not; explainability  \cite{dosilovic_explainable_2018, burkart_survey_2021}. 

The optimization that occurs in Probabilistic Graphical Model (PGM) structure learning is

\begin{align*}
    G^* = \argmax_G F(G, D) \\
    \text{subject to } G\in \mathbb{\Omega}
\end{align*}

where $D$ is a database, $G$ is a graph structure such as a BN, $F$ is a scoring function that yields the goodness of fit of the structure $G$, and $\mathbb{\Omega}$ is the set of allowed structures for $G$; for BNs this would be the space of all possible Directed Acyclic Graphs (DAGs). 

Scoring functions are essential to the structure learning problem and should have a theoretical justification in information theory or otherwise. For instance, the most common scoring functions such as Bayesian Information Criteria (BIC) \cite{schwarz_estimating_1978}, Minimum Description Length (MDL) \cite{rissanen_modeling_1978}, and Akaike Information Criterion \cite{akaike_new_1974} are all based on information theoretic criteria or can be viewed from this perspective. While we will spend part of this paper in theoretically justifying our model scoring approach, our goal is \emph{not} to present a better scoring function. Instead, our goal is to illustrate that no matter the scoring function or learning algorithm, an over-generalization is encountered when modeling at the Random Variable (RV) level.

By operating at the RV level, models force a complete distribution, as is the case with BNs. While a complete distribution is often desired, this has an unintended over-generalization consequence, particularly in under-determined domains. This phenomenon even occurs in deep learning systems, and is generally referred to as fooling \cite{szegedy_intriguing_2014, nguyen_deep_2015, kardan_fitted_2018}. However, we will limit our scope to PGMs as our end goal is to analyze and/or hypothesize structural dependency relationships. Given this goal, such over-generalization could yield non-optimal structures, biasing analysis and derived hypotheses leading to misguided conclusions. To illustrate this over-generalization and provide  intuition for learning at the RV instantiation level, we provide a motivating example taken from real-world data. 

\begin{figure*}
     \centering
     \begin{subfigure}[b]{0.2\textwidth}
         \centering
         \includegraphics[width=\textwidth]{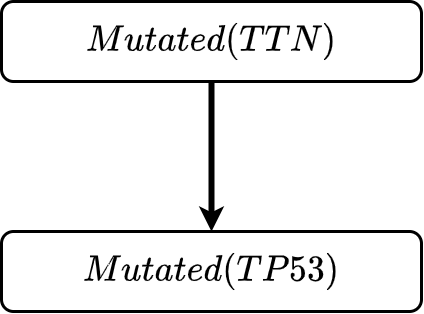}
         \caption{}
         \label{fig:bn-example}
     \end{subfigure}
     \hfill
     \begin{subfigure}[b]{0.38\textwidth}
         \centering
         \includegraphics[width=\textwidth]{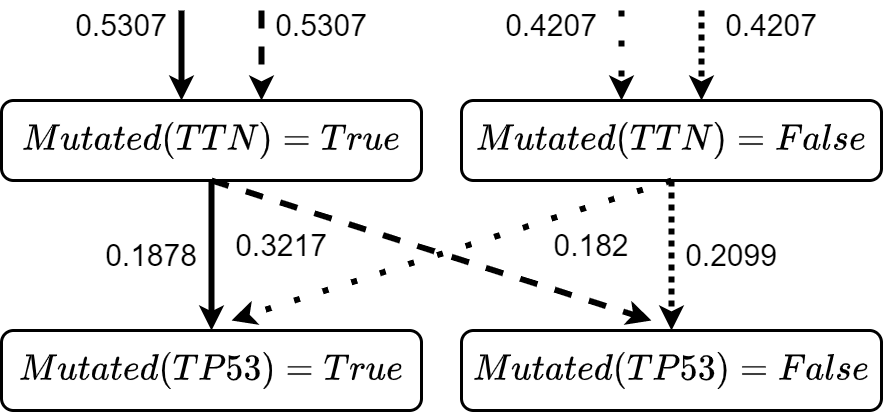}
         \caption{}
         \label{fig:bn-world-example}
     \end{subfigure}
     \hfill
     \begin{subfigure}[b]{0.38\textwidth}
         \centering
         \includegraphics[width=\textwidth]{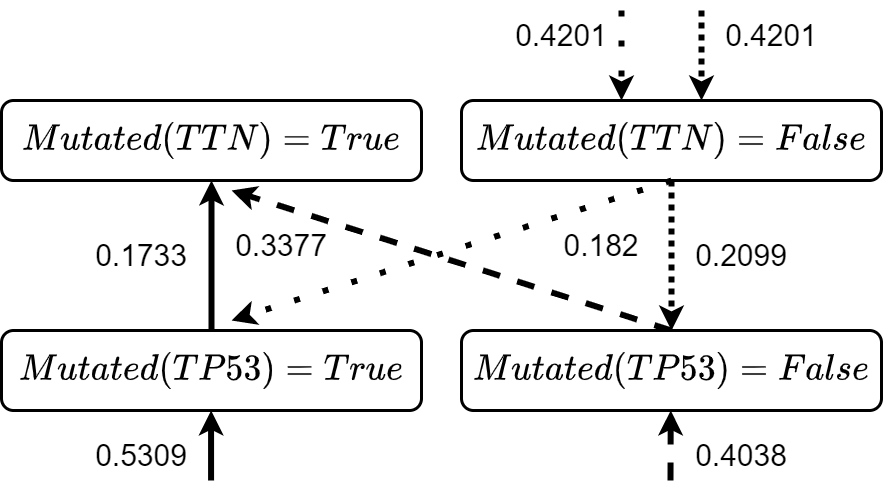}
         \caption{}
         \label{fig:bkb-example}
     \end{subfigure}
     \caption{(a) A simple learned BN over the TCGA gene mutation dataset using GOBNILP \cite{cussens_polyhedral_2015} where the variable states are either mutated or not mutated. (b) A graph of all corresponding worlds represented in (a) delineated by different line styles. (c) A better orientation of intra-world dependency relationships that lead to a lower total conditional entropy. All values are conditional entropies calculated from the TCGA gene mutational dataset.}
     \label{fig:example}
\end{figure*}

\paragraph{Motivating Example} It is well known in cancer research that the genes TP53 and TTN have somatic mutations that affect chemotherapy responses \cite{xue_ttntp53_2021}. To demonstrate a real-world effect of BN over-generalization, we learned a simple BN for this interaction over the TCGA \cite{tomczak_cancer_2015} mutational dataset as seen in Figure \ref{fig:bn-example}. This BN encodes four possible worlds represented by distinctly styled arrows in Figure \ref{fig:bn-world-example}. For this example we have reduced the state space of each gene to just mutated or not mutated. Assume our goal is to minimize the entropy or uncertainty of each world or explanation. Then the conditional entropy of the model is the sum over each world's conditional entropy which is inherently direction dependent. Since there exists many possible world edge configurations (RV instantiation dependencies), there may exist a better set of edges than those induced by the BN. Figure \ref{fig:bkb-example} shows this is true and illustrates the best collection of minimal entropy inferences for this example.

\paragraph{Contributions}
To address the over-generalization described we develop a structure learning algorithm leveraging the Bayesian Knowledge Base (BKB) framework as it inherently operates on the RV instantiation level. We accomplish this by detailing a necessary scoring metric to rank BKB models based on an MDL analysis and show theoretically that our MDL score takes over-generalization into account. Leveraging this theoretical result, we then develop our BKB Structure Learning (BKBSL) algorithm to minimize MDL and demonstrate empirically both competitive accuracy and better data fit compared to learned BNs. Further, we show that our algorithm can utilize existing optimization frameworks for DAG learning bringing BKBSL into the realm of these well studied off-the-shelf methods. Lastly, we conclude by utilizing a learned BKB to explain possible gene associations over TCGA breast cancer data. 

\section{Related Work and Preliminaries} \label{sec:related}
As the MDL principle will be our guiding force in both theoretical and empirical analysis, we provide a brief review of its applications to directed PGMs, e.g. Bayesian Networks, as these models are most applicable to our study of BKBs. Lam and Bacchus \shortcite{lam_learning_1994} first presented an MDL learning approach for BNs based on a heuristic search method seeking to spend equal time between simple and more complex BN structures. This was accomplished by extending Chow and Liu's \shortcite{chow_approximating_1968} result on recovering polytrees to general BNs via Kullback-Leibler cross entropy minimization allowing them to develop a weighting function over nodes. Their approach demonstrated that minimizing the MDL of BNs performs an intuitive trade-off between model accuracy and complexity. In their work, they also eluded to a potential subjectivity in choosing a model encoding strategy leading to research for improved MDL scores for BNs \cite{yun_improved_2004, drugan_feature_2010}. Hansen and Yu \shortcite{hansen_model_2001} detail a complete review of various MDL formulations. 

Empirical evaluation of MDL as a scoring function for BN learning has also been well studied. Yang and Chang \shortcite{yang_comparison_2002} analyzed the performance of five scoring functions: uniform prior score metric (UPSM), conditional uniform prior score metric (CUPSM), Dirichlet prior score metric (DPSM), likelihood-equivalence Bayesian Dirichlet score metric (BDeu), and MDL. They showed that MDL was able to correctly identify ground-truth network structures from a variety of possible candidates, yielding the highest discrimination ability. Liu et al \shortcite{liu_empirical_2012} also performed empirical BN learning analysis over different scoring function, namely: MDL, Akaike's information criterion (AIC), BDeu, and factorized normalized maximum likelihood (fNML). Their approach tested the recovery accuracy of each scoring method over various gold standard networks as compared to the random networks used by Yang and Chang. Their results confirm the utility of MDL as it performed best in recovering the optimal networks when sufficient data was given. 

To our knowledge there has been no work in structure learning on the RV instantiation level, likely due to the desire to learn complete distributions. Further, we have limited our comparisons to BNs as they are a predominant model in literature and provide a comparison to judge empirical results. 

\subsection{Bayesian Networks} \label{subsec:bns}
A BN is a DAG $G = (V, E)$ that represents the factorized joint probability distribution over random variables $X = (X_1, \dots, X_n)$ of the form:
\begin{equation}
    \text{Pr}(X) = \prod_i^n P(X_i \mid \Pi(X_i))
\end{equation} \label{eq:bn-distr}
such that $\Pi(X_i)$, or more concisely $\pi_i$, are the structural parents of the random variable $X_i$ according to $G$ and each node $V_i \in V$ correspond directly to a random variable $X_i$ and $n$ is the number of random variables in $X$. As the BN MDL formulation is well known in the literature, we point the reader to Appendix D.1
for a review.

\subsection{Bayesian Knowledge Bases} \label{subsec:bkbs}
Santos, Jr. and Santos \shortcite{jr_framework_1999} developed the BKB framework to generalize BNs to the random variable instantiation level and to offer knowledge engineers a semantically sound and intuitive knowledge representation. BKBs unify “if-then” rules with probability theory to provide several modeling benefits compared to BNs, fuzzy logics, etc. First, BKBs do not require complete accessibility and can even leverage incomplete information to develop potentially more representative models. Second, since BKBs operate at the instantiation level, they can handle various types of cyclic knowledge. Lastly, BKBs have both robust tuning algorithms to assist in model correction/validation \cite{santos_bayesian_2013, yakaboski_bayesian_2018} and information fusion algorithms to incorporate knowledge efficiently and soundly from disparate knowledge sources \cite{santos_fusing_2011,yakaboski_efficient_2021}. 

BKBs consist of two components: instantiation nodes (I-nodes) which represent instantiations of random variables of the form $X_i=x_{ik}$ where $k$ is the $k$-th state of $X_i$, and support nodes (S-nodes) that represent the conditional probabilities between I-node relationships of the form $X_i=x_{ik} \rightarrow q = 0.87 \rightarrow X_j=x_{jl}$. The collection of these (in)dependencies describe the BKB correlation graph. For a precise definition and a graphical depiction see Appendix D.

While BKBs can handle incompleteness and various forms of cyclicity, it is necessary that all S-nodes in a BKB obey mutual exclusivity (mutex) and associated probability semantics. Mutex guarantees that mutually exclusive events cannot be true at the same time. Concretely, we say that two sets of I-nodes, $I_1$ and $I_2$, are mutex if there exists an I-node $X_i=x_{ik_1 }\in I_1$ and $X_i=x_{ik_2}\in I_2$ such that $k_1\neq k_2$. We say that two S-nodes are mutex if the two I-node parent sets of each S-node are mutex.  

BKBs use a weight function to map S-node weights to associated conditional probabilities associated with the conditional dependence relationship described by the S-node. This weight function is analogous to the conditional probability tables (CPTs) of BNs except that BKBs do not require complete information. This weight function along with the correlation graph defines the BKB.

\paragraph{Bayesian Knowledge Base}
A Bayesian Knowledge Base (BKB) is a tuple $K=(G,w)$ where $G$ is a correlation graph $G=(I\cup S,E)$ where $I$ and $S$ are the sets of I- and S-nodes, $E \subset \{I \times S\}\cup \{S\times I\}$, and $w:S\rightarrow [0,1]$ is a weight function, such that the following properties are met:
\begin{enumerate}
    \item $\forall q\in S$, the set of all incident I-nodes to $q$, denoted $\Pi (q)$, contains at most one instantiation of each random variable.
	\item For distinct S-nodes $q_1,q_2\in S$ that support the same I-node, denoted $Head_G(q_i )$, the sets $\Pi(q_1 )$ and $\Pi(q_2)$ must be mutex.
	\item For any $Q\subseteq S$ such that (i) $Head_G (q_1 )$ and $Head_G (q_2 )$ are mutex, and (ii) $\Pi (q_1)$ and $\Pi(q_2)$ are not mutex, then $\forall q_1,q_2\in Q, \sum_{q\in Q} w(q)\leq 1$.
\end{enumerate}

The probability of an inference (world) in a BKB, denoted $\tau$, is the product of all S-node weights $w(q)$ consistent with that world (Appendix D.2).
The joint probability distribution of a BKB is the sum of all inference probabilities consistent with an evidence set $\Theta$ represented as $I_\Theta$ and given by

\begin{equation}
    P(\Theta) = \sum_{\tau\in I_\Theta}\prod_{q\in \tau} w(q)
\end{equation}

\section{The BKB Minimum Description Length} \label{sec:mdl-theory}
To construct a BKB structure learning algorithm, we first need to define a scoring function that can rank BKB structures as well as provide a theoretical justification for its utility. For these reasons we focus our attention on a Minimum Description Length (MDL) score as it is a well studied tenet of learning theory \cite{rissanen_modeling_1978}. The idea is that the best model given data should minimize both (1) the encoding length of the model and (2) the length of encoding the data given the model. This is akin to applying Occam's Razor to model selection, i.e., choose the simplest model that still describes the data reasonably well.

\subsection{Encoding the BKB}
The minimum length needed to encode a BKB is related directly to the number of S-nodes modeled by the BKB. The encoding of each S-node will contain a probability value and a parent set of I-nodes. For a problem with $n$ RVs such that each RV can have $r_i$ number of instantiations, to encode all I-nodes we would need $\log_2(m)$ number of bits where $m=\prod_i r_i$. The general BKB MDL is
\begin{equation} \label{eq:bkb-mdl}
    \sum_{q\in S}\bigg((|\Pi(q)|+1)\log_2(m) + \delta \bigg) -N\sum_\tau p_\tau \log_2(q_\tau)
\end{equation} 

where $\delta$ is the number of bits needed to store the probability value and the first term is the BKB model encoding length. From Equation \ref{eq:bkb-mdl} it is clear that we incur a modeling cost for the BKB based on the finer granularity of the model. This cost is derived in Appendix C.1.
Therefore, if we know that the distribution factorizes to a BN, there is no reason to use a BKB. However, rarely in practice do we know the true distribution of the data and in most cases the data we learn from is incomplete. This eludes to a natural rationale for using a BKB that will be theoretically justified.

\subsection{Encoding the data with a BKB}
The task of encoding dataset $D$ is centered on learning a joint distribution over random variables $X=(X_1, \dots, X_n)$. As we are focused on the discrete case, each variable $X_i$ will have $r_i$ discrete states and every unique choice of random variable instantiations defines a complete world $\tau$ of the ground-truth distribution of the data and is assigned an associated probability value, $p_\tau$.

We will make several standard statistical assumptions. We assume that each data instance in $D$ is a complete world such that each data instance specifies a value for every random variable in $X$\footnote{Only for simplicity do we make the assumption that a data instance must be complete, i.e., not contain any missing values.}. We assume that each data instance is a result of independent random trails and expect that each world would appear in the dataset with a frequency $\approx N p_\tau$. 

The main theme of the MDL metric is to efficiently compress $N$ data instances as a binary string such that we minimize the string's encoding length. There are many different coding and compression algorithms we can use, but since we simply care about comparing different MDLs as a metric we can limit our focus to just symbol codes \cite{mackay_information_2005}. Symbols codes assign a unique binary string to each world in the dataset. For instance, the world $\{X_1=1, X_2=1, X_3=0\}$ might be assigned the symbol 0001. We can then encode the dataset as just the concatenation of all these world symbols and judge our compression effectiveness by calculating the length of this final binary string.

Research in information theory has proved that for symbol codes it is possible to minimize the length of this encoded data string by leveraging the probability/frequency of each world symbol occurring. Specifically, Huffman's algorithm generates optimal Huffman codes \cite{mackay_information_2005}, i.e., optimal symbol code mappings, that yield a minimum length. The key intuition for Huffman codes is that we should give shorter code words to more frequently occurring worlds and longer ones to less probable worlds. Lam and Bacchus \shortcite{lam_learning_1994} proved that the encoding length of the data is a monotonically increasing function of the cross-entropy between the distribution defined by the model and the true distribution.

Therefore, if we have a true distribution $P$ and a model distribution $Q$ over the same set of worlds, $\tau_1, \dots, \tau_t$, where each world $\tau_i$ is assigned the probability $p_i$ by $P$ and $q_i$ by $Q$, then the cross entropy between these distributions is
\begin{equation} \label{eq:cross-entropy}
    C(P,Q) = \sum_{i=1}^tp_i\log_2\frac{p_i}{q_i} = \sum_{i=1}^tp_i(\log_2p_i - \log_2q_i)
\end{equation}

Calculating Equation \ref{eq:cross-entropy} is not appealing as the number of worlds is combinatorial in the number of variables. Chow and Lui \shortcite{chow_approximating_1968} developed a famous simplification of Equation \ref{eq:cross-entropy} as just a local computation over low-order marginals when $Q$ has a tree factorization. Lam and Bucchus \shortcite{lam_learning_1994} extended their result to models that have a  general DAG structure. Their main result concludes that $C(P,Q)$ is a monotonically decreasing function of the sum of each random variable's mutual information with their parents, $I(X_i; \Pi(X_i))$. Their exact result is restated in Appendix Theorem 3. 

Further generalizing these results to the instantiation level, we can now show that an optimally learned BKB can encode the distribution as well or better than a BN. Consider again the fundamental MDL problem of learning the dataset encoding. Equation \ref{eq:cross-entropy} says that we need to minimize the total cross-entropy between $P$ and $Q$. However, in terms of data encoding, we only need to minimize the difference between each $p_i$ and $q_i$ for unique worlds that are actually in $D$. In this sense, our database encoding doesn't care about the worlds that aren't in the dataset, but for which a model like a BN naturally defines and generalizes. Therefore, BKBs handling of incompleteness gives us an opportunity to more tightly fit the data we actually know. Consider the following cross-entropy 
\begin{align}
    C(P,Q) = \sum_{i=1}^d & p_i(\log_2p_i - \log_2q_i) \nonumber \\
    & + \sum_{i=d+1}^t p_i(\log_2p_i - \log_2q_i)
\end{align}
where worlds $\tau_1, \dots, \tau_d$ are represented by the unique exemplars in $D$ that we hope to encode, i.e., $\{\tau_i, \dots, \tau_d\} = \{d_1, d_2, d_3, \dots\}_{\neq} \subseteq D$, and worlds $\tau_{d+1}, \dots, \tau_t$ are worlds that our model induces. In terms of encoding length we can narrow our focus to only considering worlds present in $D$. As Lam and Bacchus \shortcite{lam_learning_1994} proved the encoding length of the data is a monotonically increasing function cross-entropy, it is trivial to prove the following corollary.
\begin{corollary} \label{cor:c-d-mono}
Let's Define the cross-entropy $C_D(P,Q) = \sum_{i=1}^d p_i(\log_2p_i - \log_2q_i)$ between two distributions $P$ and $Q$ over the same set of worlds $\tau_1, \dots, \tau_d$ s.t. these worlds must be included in a dataset $D$. Then the encoding length of the data $D$ is monotonically increasing function of $C_D$.
\end{corollary}

Combining Corollary \ref{cor:c-d-mono} with Lam and Bacchus mutual information theorem (Appendix D Theorem 3)
we arrive at our main theoretical result.

\begin{theorem}\label{thm:c-d-mutual-info}
$C_D(P,Q)$ is a monotonically decreasing function of
\begin{equation} \label{eq:instant-mutual-info}
    \sum_{\tau\in D_{\neq}}\sum_{i=1}^n p(x_{i\tau}, \pi_{i\tau})\log_2 \frac{p(x_{i\tau}, \pi_{i\tau})}{p(x_{i\tau})p(\pi_{i\tau})}
\end{equation}
where $x_{i\tau}$ is the instantiation of random variable $X_i$ determined by data instance $\tau$, $\pi_{i\tau}$ is the parent set instantiation of random variable $X_i$ governed by $\tau$, and $D_{\neq}$ is the set of unique data instances (worlds) represented in $D$. Therefore, $C_D(P,Q)$ will be minimized when Equation \ref{eq:instant-mutual-info} is maximized.  
\end{theorem}

We leave the proof of this theorem to Appendix B.1
as the intuition is fairly straightforward from Lam and Bacchus' theorem (Appendix D Theorem 3).
We have established that the encoding length of the data is a increasing function solely of $C_D$ and that maximizing Equation \ref{eq:instant-mutual-info} minimizes  $C_D$ and thereby the encoding length. With these results, we can deduce the existence of a theoretical BKB that will have an equal to or better data encoding length than the induced worlds of a BN given $D$.

\begin{theorem}\label{thm:bkb-better-bn}
Given a BN $G$ learned over a dataset $D$ as to maximize the weight $W_G = \sum_i I(X_i; \Pi(X_i))$ given the structure $G$, there exists a BKB $K$ with a weight $W_K$ according to Equation \ref{eq:instant-mutual-info} such that $W_K \geq W_G$.   
\end{theorem}

We defer a detailed proof of Theorem \ref{thm:bkb-better-bn} to Appendix B.2
and provide a more concise proof sketch. The key insight is that any BN $G$ can be transformed into a corresponding BKB $K_G$ as BKBs subsume BNs. We are only interested in the data encoding length which can now be calculated over $K_G$ by summing over the complete worlds represented in $G$. Consider a single random variable $X_i$ and it's associated parent set $\Pi(X_i)$. For each instantiation of $X_i$ and $\Pi(X_i)$ there will be an associated S-node created in $K_G$ with an instantiated weight according to Equation \ref{eq:instant-mutual-info}. Since the choice of parent set for each random variable instantiation in $\Pi(X_i) \cup \{X_i\}$ is governed by $G$, we don't consider other S-node configurations of the same instantiated random variable set that may have greater weight. The BN structure constrains our S-node structures. Therefore, if we start with a optimal BN, transform it into a BKB $K_G$, and analyze every permutation of each S-node's possible configurations taking the permutation that maximizes the instantiated weight, we will end up with a BKB $K$ with the same number of S-nodes that has a total weight equal to or greater than the BN representation $K_G$. This result allows us to also state the following corollary based on the fact that $I(X_i;\Pi(X_i)) = H(X_i) - H(X_i|\Pi(X_i))$. 

\begin{corollary} \label{cor:cond-entropy}
Since $I(X_i;Pi(X_i)) = H(X_i) - H(X_i|\Pi(X_i)) \geq 0$. We can maximize Equation \ref{eq:instant-mutual-info} by minimizing the instantiated conditional entropy $H(x_{i\tau}|\pi_{i\tau}) = p(x_{i\tau}, \pi_{i\tau})\log_2\frac{p(x_{i\tau}, \pi_{i\tau})}{p(\pi_{i\tau})}$.
\end{corollary}

\section{BKB Structure Learning} \label{sec:bkbsl}
Theorem \ref{thm:c-d-mutual-info} dictates that for every random variable instantiation $x_{i\tau}$ in a data instance (world) $\tau\in D_{\neq}$ where $D_{\neq}$ is the set of unique data instances in $D$, we should assign an instantiated parent set $\pi_{i\tau}$ such that the instantiated conditional entropy is minimized according to Corollary \ref{cor:cond-entropy}. The key insight of our structure learning approach is that we can decompose our learning over the worlds represented in the data. In each world, we will have at most a single instantiation of each RV and our goal is to select a set of S-nodes with a structure that minimizes instantiated conditional entropy for that world.  We can view each world in the data as a separate complete inference which form an acyclic subgraph of their respective encompassing BKB. A precise definition of a BKB inference can be found in Appendix D.

Our structure learning algorithm reduces to finding a directed acyclic inference graph for each world that minimizes $\sum_\tau\sum_{i} H(x_{i\tau}|\pi_{i\tau}) = p(x_{i\tau}, \pi_{i\tau})\log_2\frac{p(x_{i\tau}, \pi_{i\tau})}{p(\pi_{i\tau})}$. Further, we can use any off-the-shelf DAG learning algorithm to accomplish this step so far as our scoring function inherently minimizes instantiated conditional entropy and BKB encoding length. There has been significant advancements in field of BN and DAG learning and we make no attempt in covering all such procedures. Instead we will focus on the state-of-the-art exact BN (DAG) solver GOBNILP \cite{cussens_bayesian_2012, cussens_polyhedral_2015}.

Upon learning each minimal entropy inference, we then need a method for merging this knowledge together that is semantically sound. A standard union type operation will not generally be supported as the unioned BKB would likely incur many mutex violations as seen in Appendix Figure 4a.
Instead, we can employ a well-studied BKB fusion \cite{santos_fusing_2011, yakaboski_efficient_2021} algorithm that supports the fusion of an arbitrary number of BKB Fragments (BKFs) by attaching source I-nodes to every S-node corresponding to the data instance from which the inference graph originated. A graphical example of this approach is depicted in Appendix Figure 4b
along with additional information regarding BKB fusion in Appendix D.3. 
This procedure ensures that that no mutual exclusion violations are present in the fused BKB maintaining a consistent probability distribution over the data and leading to model generalization. Appendix D.3
provides a detailed explanation of generalization in fused BKBs.  

Aside from forming a mutual exclusive BKB, fusion also presents us with another degree of freedom during learning. If each data instance was generated by an i.i.d. process, it is natural to assume a normalized probability over all source nodes. However, many processes do not generate truly i.i.d or representative samples. Therefore, if we view these source S-nodes as reliabilities that can be tuned, we may be able to correct errors in higher order inference calculations that arise due to under-fitting or over-generalization. We leave such analysis to future work. Combining each of the steps presented so far, we outline our general BKB structure learning procedure in Algorithm \ref{alg:bkbsl}.

\begin{algorithm}[t]
\caption{BKB Structure Learning}\label{alg:bkbsl}
\textbf{Input}: Dataset $D$, Source Reliabilities $R$, DAG learning algorithm $f$ and hyperparameters $\Theta$\
\begin{algorithmic}[1]
\STATE $K \gets \emptyset$
\FOR{$\tau \in D_{\neq}$}
    \STATE $G_{\tau} \gets f(\tau, R, \Theta)$
    \STATE $K \gets K \cup \{G_{\tau}\}$
\ENDFOR
\STATE \textbf{return} $\text{BKB-Fusion}(K, R)$
\end{algorithmic}
\end{algorithm}

\section{Empirical Results}
To demonstrate the utility of both our proposed algorithm as well as our learned BKB models we conducted 40 experiments on benchmark datasets comparing BKBSL and BNSL in terms of MDL and complexity performance. We then conducted 22 cross validation classification experiments to compare accuracy performance with learned BNs as well as a use-case studying the under-determined bioinformatics domain of structural dependency analysis among single-nucleotide polymorphism (SNPs) in breast cancer.

\subsection{Benchmark Analysis} \label{subsec:bench}
When comparing MDL between our learned BKBs and BNs, we are only concerned with comparing the \textit{data} encoding length, as the model encoding length is only used to penalize more complex models. Our \textit{data MDL} results in Appendix Table 1
demonstrate that a BKB learned using our BKBSL algorithm finds a tighter data fit than the best BN in all 40 dataset. Intuitively, this is because the BN must generalize away from potentially good instantiated scores in favor of the entire random variable score.

Our MDL experiments also demonstrates a practical strength of BKBSL over BNSL related to the number of calls to a joint probability calculator or estimator. In order to calculate the necessary scores for an exact DAG learning algorithm like GOBNILP, we needed to calculate empirical joint probabilities from each dataset. For all experiments we tracked the number of unique calls to this function by our BKBSL algorithm and traditional BNSL algorithm. Since BNSL operates at the RV level, it had to calculate all joint probabilities governed by a given parent set configuration. However, BKBSL did not need to calculate the full CPTs as it operates at the RV instantiation level and decomposes over each data instance, reducing the number of calls to this calculator. We feel that this is a more representative complexity performance metric as it is agnostic to equipment configurations. This effect is detailed in Appendix Table 1,
and we can see strong correlation between performance savings over BNs and the number of features (Pearson $r=-0.5994$, $p$-value $= \num{4.363e-5}$) as well as the number of I-nodes ($r=-0.4916$, $p$-value $= 0.0013$) in the dataset. All learned BKBs and BNs are hosted on Github and can be viewed within in a Jupyter Notebook for easier visualization. 

To finalize our BKBSL benchmark analysis we performed accuracy comparisons between our learned BKBs and traditionally learned BNs using GOBNILP and standard MDL scoring. We performed a 10-fold classification cross validation on a subset of only 22 datasets due to the increased learning and reasoning time incurred by running cross validation analysis. We can see from Appendix Table 2
that our BKBSL algorithm is very competitive with BNs in terms of precision, recall and F1-score. Further, our BKBSL models even beat BNs in $63\%$ of cases in terms of precision with greater degradation of performance in terms of recall and F1-score. The alternative hypothesis that either traditionally learned BNs or our learned BKBs will outperform each other in all accuracy categories (Chi$^2$ Statistic $\chi = 1.0$, $p$-value $= 0.3173$) is \emph{not} statistically significant. Therefore, we fail to reject the null that learned BNs or BKBs perform better in these cases owing to approximately equal total performance.

This raises the question: Why does our learned BKB perform better in some datasets and not in others? While no feature of the datasets provided any statistically significant predictor of superior performance and leaving more in-depth analysis to future work, we do hypothesize an explanation. It is a well-known problem that real world datasets are often unfaithful to DAGs, e.g. BNs, due to the existence of multiple information equivalent Markov Boundaries (MBs) \cite{statnikov_algorithms_2013, wang_causal_2020}. Since our BKBSL focuses on learning an optimal structure for every unique exemplar $\tau$, we can view each learned BKF as an equivalent inference from a hypothetical BN whose dependency structure matches that of the associated BKF. As we are only concerned with the specific instantiations of $\tau$, the hypothetical BN and BKF will yield the same probability for this world as their parameters are governed by the same dataset. As our prediction task is to determine the most likely state of a response variable $Y$ given a complete set of evidence $E$, e.g., $y^* = \argmax_y Q(Y=y \mid E)$, then the closer our joint probability $Q(Y=y, E)$ is to the true data distribution $P(Y=y, E)$ the more accurate our classifier. This is due to the fact when comparing all conditional probabilities $Q(Y=y_i \mid E) = \frac{Q(Y=y_i, E)}{Q(E)}$ the denominator cancels out and we are only concerned with the accuracy of $Q(Y=y_i, E)$. If we imagine our learned BKFs deriving from various hypothetical BNs each with uniquely induced MBs for every RV, our fused BKB essentially incorporates a multiplicity of MBs choices for each of these hypothetical BNs and selects the best performing choice for every world of interest, i.e., prediction class given evidence. We hypothesize that our BKBSL will then perform better on datasets that induce more information equivalent MBs since a BN must select only one and our BKBSL can incorporate multiple in its predictions. Whereas in datasets with fewer MBs, our BKBSL performance may degrade due to overfitting. We intend to study this area further as it may yield clear indications about when to use BKBSL over BNSL in practice.

\subsection{Gene Regulatory Network (GRN) Application in Breast Cancer}
We applied our approach to somatic mutation profiles of breast cancer cases in TCGA \cite{tomczak_cancer_2015} to study whether our learned model could still offer utility in this extremely under-determined domain. Since prediction accuracy would not be a reliable metric of success in this dataset, we focused our analysis on hypothesizing potentially significant mutational interactions in cancer. However, if we are to trust any structural hypotheses generated by our approach, we need to ensure the model captures two fundamental biological concepts: (1) We can extract two- or three-hit interactions that are supported in the literature \cite{knudson_mutation_1971, knudson_two_2001, segditsas_apc_2009}, and (2) we can identify and (possibly) handle genomic instability \cite{bai_pik3ca_2014, croessmann_pik3ca_2017}. 

\begin{figure*}
     \centering
     \begin{subfigure}[b]{0.4\textwidth}
         \centering
         \includegraphics[width=\textwidth]{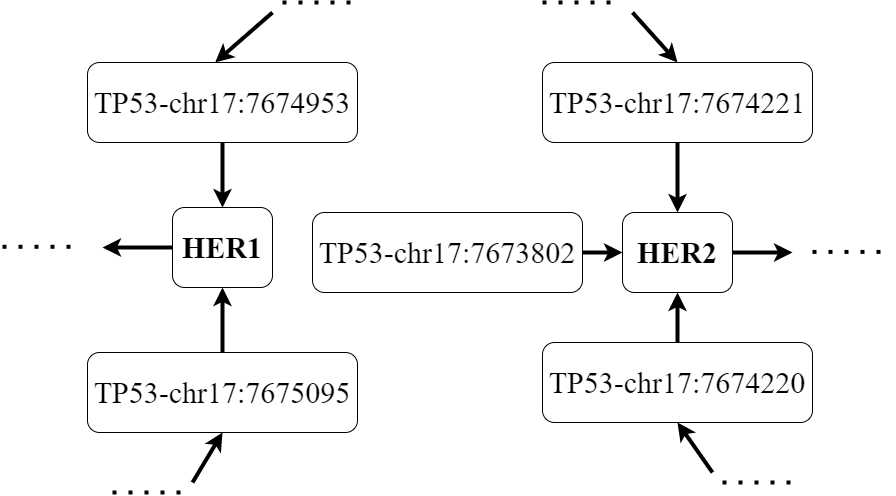}
         \caption{}
         \label{fig:tp53-her}
     \end{subfigure}
     \hfill
     \begin{subfigure}[b]{0.48\textwidth}
         \centering
         \includegraphics[width=\textwidth]{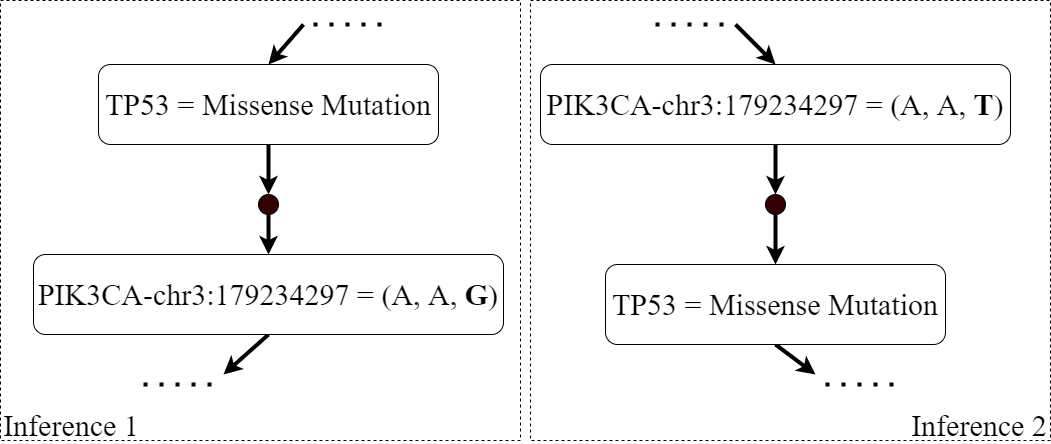}
         \caption{}
         \label{fig:tp53-pik}
     \end{subfigure}
     \caption{(a) RV level graph of learned BKB over TCGA breast cancer subgraphed on TP53 SNP relationships with HER genes. (b) Genomic instability evidence from the RV level cycle between PIK3CA SNP and general TP53 gene variable. To capture multiple levels of granularity, we included features related to exact positional SNPs as well as including gene level features and general variant classifications. For details about our naming conventions and feature selection process see Appendix A.
     }
     \label{fig:gene}
\end{figure*}

Given the well-regarded two- and three-hit hypotheses for understanding the role of genetic mutations in cancer development, a model attempting to describe a mutational GRN should be able to capture this concept. The premise behind the two- or three-hit hypotheses is that because many cancers are driven by mutations in various tumor suppressor genes and these loss-of-function mutations are recessive in nature, in general, at least two mutations in these genes are needed to develop cancer. There are certain tumor suppressors genes such as TP53 \cite{kandoth_mutational_2013, muller_mutant_2014} that are common in all cancer sub-types and are likely the first hit for non-hereditary cancers. Looking at the dependence relationship subgraph in our learned BKB between a first hit tumor suppressor gene such as TP53 and a second hit tumor suppressor gene related to breast cancer such as a HER1 or HER2 \cite{osborne_oncogenes_2004}, we should observe a directed dependency relationship from TP53 to HER1 or HER2. Figure \ref{fig:tp53-her} shows we observe this relationship adding to the biological validity of our model.

It is well known that cancer is also driven by genomic instability \cite{negrini_genomic_2010}, the increased tendency for gene mutations to accumulate, disrupting various biological pathways and in turn causing more mutations. An artifact that we discovered in our BKB network analysis and illustrative of our first example is that in some cases there exist cyclic relationships on the random variable level. This is due to separate inferences learning opposing dependency relationships due to the additional context of their respective inference. This effect can be seen in Figure \ref{fig:tp53-pik}. Here we have a cyclic random variable relationship between TP53 and the SNP located at chromosome 17 position 179234297 of the PIK3CA gene. TP53 and PIK3CA are known drivers of genomic instability \cite{bai_pik3ca_2014}, affecting the mutational states of many other genes in their inference contexts. Since our inference level parent set limit was set to one, our algorithm cannot reliably extract mutual dependency relationship between TP53 and PIK3CA, thereby causing different inferences to have different directionalities. This cannot be captured by a BN. Further, this result is supported by in-vitro research regarding the joint effect of PIK3CA and TP53 on genomic instability in breast cancer \cite{croessmann_pik3ca_2017} and is expected from this model given that both genes drive many other downstream mutations. Overall we found 12 of these cyclic relationships in our learned BKB. Of these 12 associations we found literature support for four of them, namely, PIK3CA and TP53 \cite{croessmann_pik3ca_2017}, OBSCN and TP53 \cite{sjoblom_consensus_2006, perry_obscurins_2013}, MAP3K1 and PIK3CA \cite{avivar-valderas_functional_2018}, CAD and PIK3CA \cite{wu_activation_2014}. 

\section{Limitations and Future Work} \label{sec:limits}
The primary limitation of our approach is that we need to learn a BKB fragment for every instance in the training dataset. While we have some complexity benefits from making fewer calls to a given joint probability calculator, this benefit could be neutralized due to the requirement of having to run the DAG learning algorithm multiple times. For instance, once all scores are calculated, BNSL only requires a single run of the respective DAG learning algorithm, whereas our BKBSL approach requires running this algorithm $N$ times. We leave to future work the exploration of this trade off while hypothesizing that there may be a DAG learning formulation in which all fragments can be learned in a single pass by reusing/sharing local scores/structures between fragments. 

While our BKBSL algorithm seems to generalize well to unobserved examples in our benchmark experiments, we still see instances with significant accuracy degradation. It is a largely unanswered question as to why machine learning algorithms perform better or worse on particular datasets \cite{pham_impact_2008} and we have detailed a possible explanation in Section \ref{subsec:bench} to be explored in future work. As mentioned in Section \ref{sec:bkbsl}, we could also address accuracy degradation by tuning the source node reliabilities in our model. Such an approach yields another degree of freedom to adjust the model and also may highlight the importance/significance of individual data instances in relation to over model accuracy. We also leave this direction for future research.

\section{Conclusions}
We have presented a new approach for performing Bayesian structure learning at the random variable instantiation level by leveraging Bayesian Knowledge Bases. We have detailed a theoretical justification for our algorithm and learned model as being the fusion of minimal entropy inferences or explanations over each training exemplar. We demonstrated empirically that our BKBSL algorithm finds a superior BKB than an equivalent BN (scored as BKB) on 40 benchmark datasets using our MDL formulation. Further, we demonstrated the practical utility of our approach by presenting statistically competitive accuracy with learned BNs over 22 benchmark datasets using a 10-fold cross validation. This provides evidence that our algorithm adequately generalizes to unseen data based on known knowledge rather than over-generalizing to force a complete distribution. Lastly, we conducted a structural analysis over a gene regulatory network learned from breast cancer mutation data taken from TCGA. This analysis resulted in finding dependency relationships that matched biological intuition and also revealed associations that are well known in the bioinformatics community.  

\section*{Acknowledgements}
This research was funded in part by the National Center for Advancing Translational Sciences (NCATS), National Institutes of Health, through the Biomedical Data Translator program (NIH award OT2TR003436), Air Force Office of Scientific Research Grant No. FA9550-20-1-0032 and DURIP Grant No. N00014-15-1-2514. Special thanks to Joseph Gormley and Eugene Hinderer for their comments and encouragement.

\bibliography{references.bib}

\onecolumn
\section*{Appendices}
\input{appendix}

\end{document}

%% file: appendix.tex
\appendix

\section{Experimental Results and Design} \label{app:experiments}
\input{Tables/keel-mdl}
\input{Tables/keel-accuracy}
\input{Tables/keel-info}

\subsection{Benchmark Experiments}
All benchmark datasets were taken directly from the KEEL data repository \cite{alcala-fdez_keel_2011} and contained no missing values in order to facilitate direct BNSL comparisons. All preprocessed KEEL datasets used in our analyses are hosted on Zenodo\footnote{\url{https://zenodo.org/record/6584753#.Yo_PK1RBxPZ}}. All 10-fold cross validation experiments analysis used splits already implemented in the KEEL repository for each dataset. All experiments were performed on cluster of 15 Dell PowerEdge M630 blade servers each with 32 processors and 500GB of RAM and leveraging the Ray \cite{moritz_ray_2018} library to distribute all calculations. While distributed, we still utilized non-optimized python code for all GOBNILP score calculations limiting the number of parent sets that could be explored, particularly on larger problems. In order to arrive at each parent set limit in Table \ref{tab:keel-info}, we iterated parent set limits starting from one to $n_i$ where $n_i$ is the parent set limit for each dataset $i$ that was able to complete all necessary score calculations within 12 hours. For all MDL and accuracy analysis we learned BKBs and BNs on their respective highest parent set limit $n_i$ scores calculated for each dataset in order to explore the greatest possible DAG space available. We calculated all joint probabilities based on raw dataset counts without employing any estimation techniques. All continuous valued features where discretized into at least two splits using the Entropy MDL method \cite{fayyad_multi-interval_1993}.

\subsection{TCGA Breast Cancer Experiments}
As there are over 500 features in our TCGA dataset we will not list them here and instead point the reader to the pre-processed dataset available on Zenodo\footnote{\url{https://zenodo.org/record/6584753#.Yo_PK1RBxPZ}}. Features annotated as \{Gene Name\}-\{Chromosome \#\}-\{Position \#\} are SNP mutations with states represented as a tuple (Reference Allele, Tumor Allele 1, Tumor Allele 2) and features represented as just gene names (HGNC Symbols\cite{tweedie_genenames_2021}) have states corresponding to the variant mutation classification that occurred anywhere in that gene, i.e., missense mutation, nonsense mutation, etc. 

As many mutations only occurred in a single case, we used a feature selection process to filter out many mutations that would not have been statistically significant. Our feature selection process involved a threshold for how many cases a particular mutation and state appeared. For the SNP features we set an extremely low inclusion threshold of 2, resulting in 72 unique SNP features, and for whole gene features we set a threshold of 10, resulting in 428 features. Our filtered dataset contained 939 breast cancer cases and our parent set limit was set to 1 in order to control complexity for analysis.

\section{Proofs} \label{sec:proofs}
\subsection{Theorem \ref{thm:c-d-mutual-info}} \label{proof:cd-mutual-info}
$C_D(P,Q)$ is a monotonically decreasing function of
\begin{equation} \label{eq:cd-mutual-info-appendix}
    \sum_{\tau\in D_{\neq}}\sum_{i=1}^n p(x_{i\tau}, \pi_{i\tau})\log_2 \frac{p(x_{i\tau}, \pi_{i\tau})}{p(x_{i\tau})p(\pi_{i\tau})}
\end{equation}
where $x_{i\tau}$ is the instantiation of random variable $X_i$ determined by data instance $\tau$ and $\pi_{i\tau}$ is the parent set instantiation of random variable $X_i$ governed by $\tau$.
\begin{proof}
We can write $C_D(P, Q)=\sum_{\tau \in D_{\neq}}C_\tau(P, Q)$ where 
\begin{align}
    C_\tau(P,Q) &= p_\tau\log p_\tau - p_\tau\log q_\tau \\
    &= p_\tau\log p_\tau - p_\tau \sum_{i=1}^n \log p(x_{i\tau} \mid \pi_{i\tau}) \\
    &= p_\tau\log p_\tau - p_\tau\sum_{i=1}^n \log\bigg(\frac{p(x_{i\tau}, \pi_{i\tau})}{p(x_{i\tau})p(\pi_{i\tau})}\bigg) - p_\tau\sum_{i=1}^n \log p(x_{i\tau})
\end{align}
Now, consider that $p(x_{i\tau}) = p_\tau + \tilde{p}(x_{i\tau})$, such that $\tilde{p}(x_{i\tau}) = P(X_1\neq x_{1\tau}, \dots, X_i=x_{i\tau}, \dots, X_n\neq x_{n\tau})$ and similarly, $p(x_{i\tau}, \pi_{i\tau}) = p_\tau + \tilde{p}(x_{i\tau}, \pi_{i\tau})$. Then,

\begin{equation}
     -p_\tau\sum_{i=1}^n \log p(x_{ij}) = -\sum_{i=1}^n p(x_{i\tau})\log p(x_{i\tau}) +\sum_{i=1}^n \tilde{p}(x_{i\tau})\log\tilde{p}(x_{i\tau})
\end{equation}
and 
\begin{align}
    - p_\tau\sum_{i=1}^n \log\bigg(\frac{p(x_{i\tau}, \pi_{i\tau})}{p(x_{i\tau})p(\pi_{i\tau})}\bigg) = -\sum_{i=1}^n &p(x_{i\tau}, \pi_{i_\tau}) \log\bigg(\frac{p(x_{i\tau}, \pi_{i\tau})}{p(x_{i\tau})p(\pi_{i\tau})}\bigg) \\
    &+ \sum_{i=1}^n \tilde{p}(x_{i\tau}, \pi_{i_\tau}) \log\bigg(\frac{p(x_{i\tau}, \pi_{i\tau})}{p(x_{i\tau})p(\pi_{i\tau})}\bigg)
\end{align}
Further we can define the instantiated entropy as $H_1(x_{ij}) = -p(x_{ij})\log p(x_{ij})$ to yield
\begin{align}
    C_\tau(P,Q) = -\sum_{i=1}^n &p(x_{i\tau}, \pi_{i_\tau}) \log\bigg(\frac{p(x_{i\tau}, \pi_{i\tau})}{p(x_{i\tau})p(\pi_{i\tau})}\bigg)
    + \sum_{i=1}^n \tilde{p}(x_{i\tau}, \pi_{i_\tau}) \log\bigg(\frac{p(x_{i\tau}, \pi_{i\tau})}{p(x_{i\tau})p(\pi_{i\tau})}\bigg) \\
    &+ \sum_{i=1}^nH_1(x_{i\tau}) +\sum_{i=1}^n \tilde{p}(x_{i\tau})\log\tilde{p}(x_{i\tau}) - H_1(\tau)
\end{align}
The last three terms are functions of the data and not the underlying structure of the model and thereby a constant when comparing networks. In all worlds, it must be the case that $p(x_{i\tau}, \pi_{i_\tau}) \geq \tilde{p}(x_{i\tau}, \pi_{i_\tau})$. Therefore, $C_D(P,Q)$ is a monotonically decreasing function of Equation \ref{eq:cd-mutual-info-appendix}.
\end{proof}

\subsection{Theorem \ref{thm:bkb-better-bn}} \label{proof:bkb-better-bn}
Given a BN $G$ learned over a dataset $D$ as to maximize the weight $W_G$ given by Equation \ref{eq:sum-mutual-info} and corresponding weight $W_G'$ given by Equation \ref{eq:instant-mutual-info}, there exists a BKB $K$ with a weight $W_K$ according to Equation \ref{eq:instant-mutual-info} such that $W_K \geq W_G'$.  
\begin{proof}
Assume we have a BN $G$ with weight $W_G$ according to Equation \ref{eq:sum-mutual-info} and $W_G'$ according to Equation \ref{eq:instant-mutual-info}. We can transform $G$ into a BKB $K_G$ such that an S-node is created corresponding to every entry in $G$'s conditional probability table (CPT). To complete these proofs we need to leverage a few definitions.

\paragraph{Definition \ref{proof:bkb-better-bn}.1} A set of S-nodes $B_i$ is said to be a causal rule set (CRS) for the random variable $X_i$ if $B_i$ contains all S-nodes pointing to the instantiations of $X_i$. 

\paragraph{Definition \ref{proof:bkb-better-bn}.2} For every set of I-nodes $\alpha$ over unique random variables there is a set of possible S-nodes $\psi$ such that every S-node $q \in \psi$ corresponds to a permutation over $\alpha$. In other worlds, this set is all the unique S-nodes that can be formed from $\alpha$ using all instantiations in $\alpha$. Therefore, the cardinality of $|\psi| = |\alpha|$. 

For the BKB $K_G$ the random variables associated with the parents of every S-node in each CRS $B_i$ will be $\Pi_G(X_i)$. Since a weight according to Equation \ref{eq:instant-mutual-info} is assigned to every S-node, we can also view each CRS as having a weight $W(B_i)=\sum_{q\in B_i}W(q)$. 

\paragraph{Proposition} Starting from the BKB $K_G$ yielding the CRSs $B_1, B_2, \dots, B_N$ we can find a new set of CRSs $B_1', B_2', \dots, B_N'$ that have a total weight $\geq W(K_G)$.

Now our proof reduces to proving the above proposition. Consider an S-node $q \in B_i$ and the associated set of I-nodes $\alpha$ corresponding to $q$. We know from the above definitions that there is a set $\psi$ of all possible S-node permutations of the I-node set $\alpha$. Therefore, there must exist an S-node $q' \in \psi$ such that $W(q') \geq W(q)$. In the case that $q' \neq q$, we can remove $q$ from $B_i$ and add $q'$ to the CRS associated with $Head(q')$ which will be different from $Head(q)$. Performing this action will increase the weight of the BKB by a $\Delta(W(q), W(q')) \geq 0$. Therefore, we will always find a new BKB from a BN that has a weight equal to or greater than that of the original BN.  
\end{proof}

\section{Derivations}
\subsection{Derivation of BKB Modeling Cost} \label{der:bkb-cost}
First, let's define model MDL, denoted MMDL, for BKBs and BNs respectively
\begin{equation}
    MMDL(G_{BN}) = \sum_{i=1}^n\bigg(|\Pi(X_i)|\log(n) + \delta(r_i - 1)\prod_{X_j\in\Pi(X_i)}r_j\bigg)
\end{equation}
\begin{equation}
    MMDL(G_{BKB}) = \sum_{i=1}^n\sum_{q\in CRS(X_i)} \bigg(|\Pi(q)|\log(m) + \delta \bigg)
\end{equation}
where $CRS(X_i)$ denotes the Causal Rule Set for random variable $X_i$ and all logarithms in this derivation are base 2. Further, we know that $G_{BKB}$ will be complete and is constructed from the $G_{BN}$, and thereby $\forall q \in CVS(X_i) \rightarrow K=|\Pi_{BN}(X_i)|=|\Pi_{BKB}(q)|$. Since we know that $|\Pi_{BKB}(q)|$ will be a constant for all $q$'s in the same CRS, then we can simply the summation and just multiply by the number of different parent set instantiations of $X_i$. Therefore, let's define $R=\prod_{X_j\in\Pi(X_i)}r_j$ yielding the simplified expressions
\begin{equation}
    MMDL(G_{BN}) = \sum_{i=1}^n\bigg(K\log(n) + \delta(r_i - 1)R\bigg)
\end{equation}
\begin{equation}
    MMDL(G_{BKB}) = \sum_{i=1}^n \bigg(K\log(m) + \delta \bigg)R
\end{equation}
Then the cost for converting the BN to a complete BKB is 
\begin{align}
    C(G_{BKB}, G_{BN}) &= MMDL(G_{BKB}) - MMDL(G_{BN}) \\
    &= \sum_{i=1}^n \bigg(K\log(m) + \delta \bigg)R - \bigg(K\log(n) + (2-r_i)\delta R\bigg) \\
    &= \sum_{i=1}^n K\big(\log(m)R - \log(n)\big) + (2-r_i) \delta R \\
    &= \sum_{i=1}^n K\log\bigg(\frac{m^R}{n}\bigg) + (2-r_i)\delta R
\end{align}
Now let's define $m'=r_{max}^n$ where $r_{max}$ the the maximum number of instantiations of any variable in $X$. This yields
\begin{align}
    C(G_{BKB}, G_{BN}) &\leq \sum_{i=1}^n K\log\bigg(\frac{m'^R}{n}\bigg) + (2-r_i)\delta R \\
    &= \sum_{i=1}^n  K\big(Rn\log(r_{max}) - \log(n)\big) + (2-r_i)\delta R \\
    &= \sum_{i=1}^n  |\Pi(X_i)|\big(Rn\log(r_{max}) - \log(n)\big) + (2-r_i)\delta R 
\end{align}

\section{Other Background} \label{sec:other-background}
\subsection{Bayesian Networks} \label{app:bns}
A BN is a Directed Acyclic Graph (DAG) $G = (V, E)$ that represents the factorized joint probability distribution over random variables $X = (X_1, \dots, X_n)$ of the form:
\begin{equation}
    \text{Pr}(X) = \prod_i^n P(X_i \mid \Pi(X_i))
\end{equation} \label{eq:app-bn-distr}
such that $\Pi(X_i)$, or more concisely $\pi_i$, are the parents of the random variable $X_i$ according to $G$ and each node $V_i \in V$ correspond directly to each random variable $X_i$.

If a dataset $D$ and the BN structure $G$ are given, then parameters of a BN, or conditional probabilities, can be readily calculated via Maximum Likelihood Estimation (MLE) and resolve to simply taking counts over the dataset \cite{bouckaert_bayesian_1995}, yielding the log-likelihood:
\begin{equation}
    LL_D (G)=\sum_{i=1}^n\sum_{j=1}^{c_i}\sum_{k=1}^{r_i}N_{ijk} \log \bigg(\frac{N_{ijk}}{N_{ij}}\bigg)
\end{equation} 
where $r_i$ is the number of instantiations available to $X_i$, and $c_i = \prod_{u_{j\in \pi_i }} r_j$  which is the number of possible parent set $\pi_i$ configurations. This leads to the two predominant formulations of the BN MDL from Lam and Bacchus \cite{lam_learning_1994} and Suzuki \cite{suzuki_learning_1999, suzuki_construction_1993}, namely

\begin{equation}
    \sum_{i=1}^n[k_i\log_2(n) + \delta (r_i - 1)c_i] - N\sum_\tau p_\tau\log_2q_\tau \leq H(D|G) + \frac{K(G)}{2}\log_2(N) 
\end{equation}

where the left side of the inequality describes the MDL characterized by Lam and Bacchus and the right corresponds to the derivation by Suzuki. Further, $k_i$ is the number of parents of $X_i$, $\delta$ is the number of bits required to encode a probability value, $\tau$ is an atomic world encoded by the distribution $\text{Pr}(\cdot)$, $H(D|G)$ is the conditional empirical entropy of the data given the model $G$, and $K(G)$ is the number of parameters in the BN which is equal to the total number of conditional probability values. 

We can view both of these formulations as equivalent but taking different perspectives. When calculating the model encoding term, Lam and Bucchus's formulation derives from the arithmetic coding literature and Suzuki leverages a "universal" or Lempel-Ziv coding paradigm \cite{mackay_information_2005}. 

\subsection{Bayesian Knowledge Bases}

\begin{figure}
    \centering
    \includegraphics[width=0.5\textwidth]{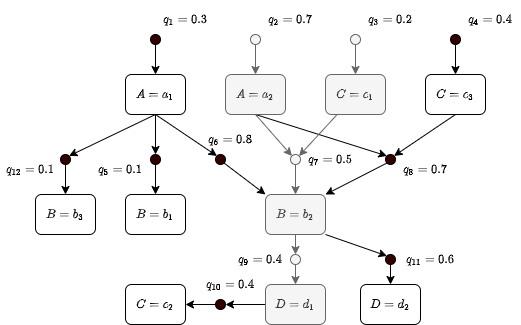}
    \caption{Example of a BKB. The greyed I-nodes and S-nodes represent an example inference (world) $\tau$ in the BKB.}
    \label{fig:bkb-example-app}
\end{figure}

In this section we will provide some useful definitions for better understanding BKBs as well as a description of BKB fusion. A graphical depiction of an example BKB can be seen in Figure ~\ref{fig:bkb-example-app}. In this example we can see the way in which the conditional (in)dependency relationships are defined on the random variable instantiation level as opposed to the random variable level. Figure ~\ref{fig:bkb-example-app} exemplifies how BKBs can encode cyclic dependencies on the random variable level such as between random variables $C$ and $D$ as well as forming an incomplete distribution in which all conditional probabilities between random variables need not be specified. 

\subsubsection{Definitions} \label{app:bkb-defs}

\paragraph{Correlation Graph} A correlation-graph is a directed graph $G=(I\cup S,E)$ s.t. $I\cap S=\emptyset$, $ E=\{I\times S\}\cup \{I\times S\}$, and $\forall q\in S \exists \alpha \in I$, such that $(q,\alpha)\in E$, where $\alpha \in I$ represent the I-nodes and $q\in S$ represent the S-nodes. 

\paragraph{BKB Inference} An inference is a tuple $\tau=(I'\cup S',E')$ which forms a subgraph over $K$ where $I'\subseteq I$, $S'\subseteq S$, and $E'\subseteq E$ and has a weight $w(\tau)=\prod_{a\in S'}w(q)$  that has the following properties:
\begin{enumerate}
    \item Every I-node $\alpha\in I'$ is well-supported s.t. $\forall \alpha$ $\exists$ $(q,\alpha)in E'$, i.e., all I-nodes have an incoming S-node,
	\item Every S-node $q\in S'$is well-founded s.t. $\forall (\alpha ,q)\in E,(\alpha ,q)\in E'$, i.e., all incoming edges to an S-node included in $\tau$ have all incoming edges associated with governing correlation graph,
	\item All S-nodes are well-defined s.t. $\forall q$ $\exists$ $(q,\alpha )\in E'$, i.e., there are no S-nodes without a head I-node,
	\item The subgraph representing $\tau$ is acyclic,
	\item Each random variable has at most one instantiation.
\end{enumerate}

\subsection{BKB Fusion} \label{subsec:bkb-fusion}
\begin{figure}[t]
     \centering
     \begin{subfigure}[b]{0.3\textwidth}
         \centering
         \includegraphics[width=\textwidth]{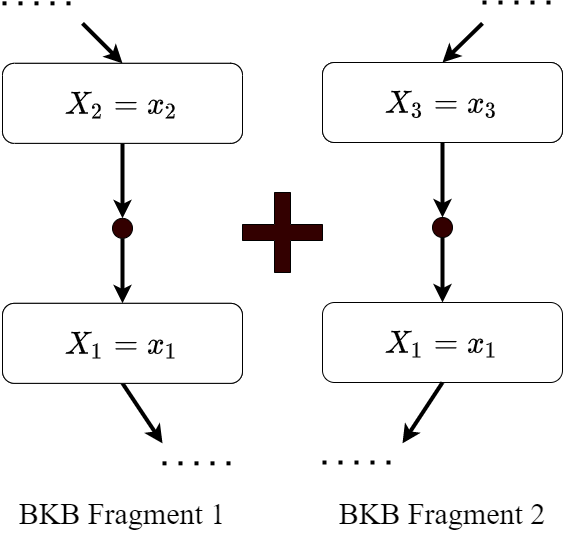}
         \caption{}
         \label{fig:frags-example}
     \end{subfigure}
     \hfill
     \begin{subfigure}[b]{0.6\textwidth}
         \centering
         \includegraphics[width=\textwidth]{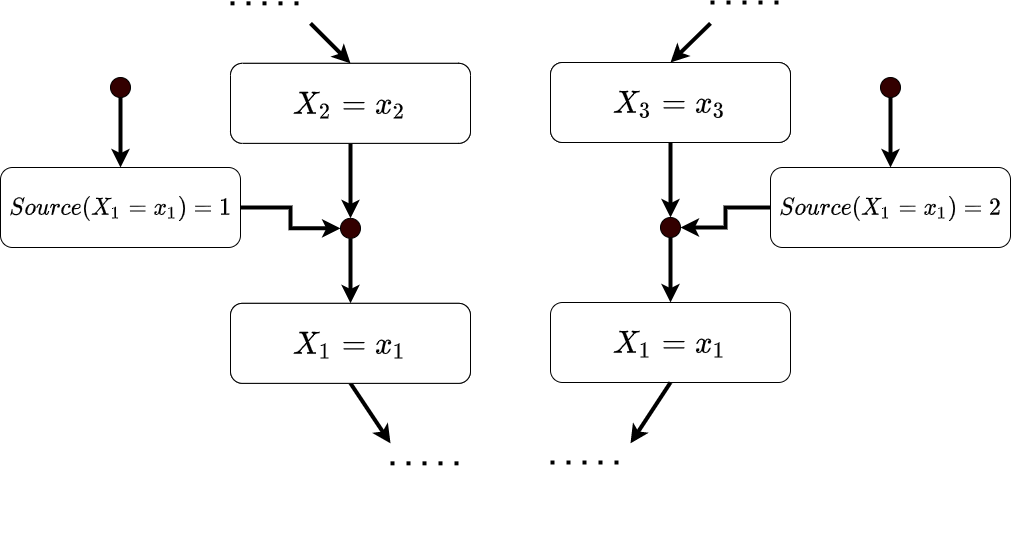}
         \caption{}
         \label{fig:fusion-example}
     \end{subfigure}
     \hfill
     \caption{(a) Example of two BKB fragments that violate the mutual exclusion property of a BKB. This can be seen by viewing the parent set of each S-node incident on the $X_1=x_1$ and noticing that these two sets are not mutex. Therefore, a simple union of these S-nodes would not be possible as it would violate our BKB semantics. (b) The BKB fusion solution that creates two distinct source I-nodes for $X_1=x_1$ thereby making the two parent sets mutually exclusive.}
     \label{fig:fusion}
\end{figure}

The core idea behind BKB fusion is to take the union of all input BKB Fragments (BKFs), which are also valid BKBs, and attach a source I-node to every incident S-node on random variable $X_i$, denoted $S_{X_i}$, with an instantiation equaling the designation of the source BKF, denoted $\sigma$, from which the S-node originated. Further, BKB fusion takes into account the reliability or probability of correctness of each source such that more correct or reliable sources will have greater weight. Intuitively, adding source I-nodes in this way transforms a rule described by an S-node such as ``If $A$ is true, then $B$ is true with probability $p$'' into ``If source $\sigma$ is correct and $A$ is true, then $B$ is true with probability $p$.'' A graphical depiction of this procedure can be seen in Appendix Figure \ref{fig:fusion-example} and an exact algorithm follows.

Given a set of $n$ BKFs $\{K_1, K_2, \dots, K_n\}$ where $K_i = (G_i, w_i, \sigma_i, r(\sigma_i))$, $G_i=(I_i, S_i, E_i)$, $\sigma_i$ is the name of the BKF source, and $r(\sigma_i)$ is the reliability index of that source. Then the output of Algorithm \ref{alg:bkb-fusion} is a new BKB, $K'=(G', w')$ with $G'=(I', S', E')$ that is a fusion of these fragments. For an I-node $\alpha$ in some fragment, let us denote $R_\alpha$ as the random variable corresponding to the instantiation $\alpha$. For a source node $s=(S_{R_\alpha} = \sigma_i)$ let the reliability of that $s$ be $r(s) = r(\sigma_i)$. A more detailed discussion can be found in the work of Santos et al \cite{santos_fusing_2011}. 

\begin{figure}
    \centering
    \includegraphics[width=0.7\textwidth]{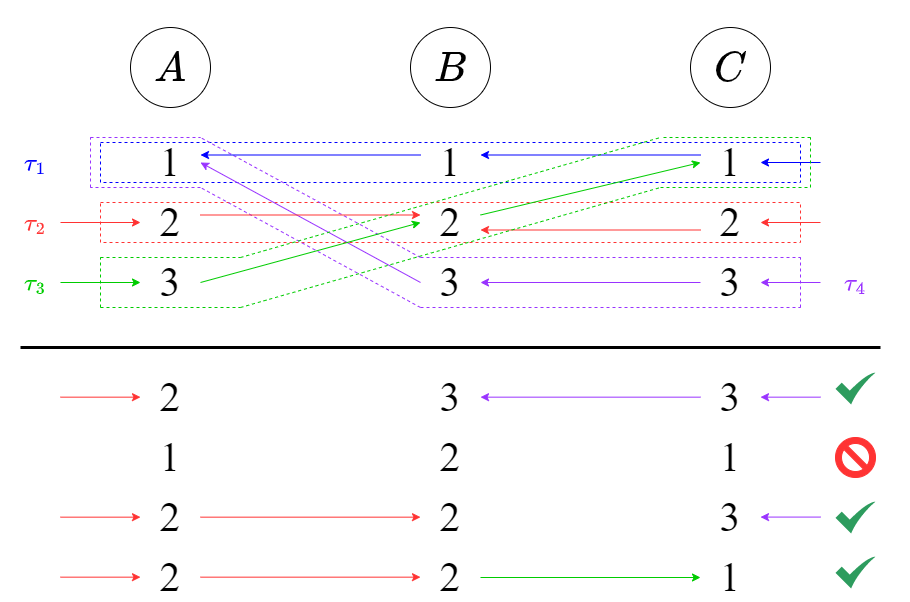}
    \caption{An image depicting BKB generalization through the fusion of four BKB fragments learned in the context of four different data exemplars. The circles above represent the random variables $A$, $B$, and $C$ each having 3 states. The four learned BKFs are denoted $\tau_i$ and the dependency structure for each BKF in $\tau_i$ is designated by their respective colored arrows. The bottom portion of the figure shows four other world instantations that could be observed in unseen data and it is clear that our BKB can generalize with fusion to three of the four unseen worlds.}
    \label{fig:bkb-generalize}
\end{figure}

\paragraph{BKB Generalization through Fusion} To better clarify the mechanism by which BKB fusion allows our BKBSL algorithm to generalize to unseen data we leverage the visualization in Figure ~\ref{fig:bkb-generalize}. Figure ~\ref{fig:bkb-generalize} depicts an example collection of three random variables, $A$, $B$, and $C$ each having three discrete states 1, 2, and 3. This random variable specification yields a total of $3^3=27$ unique worlds that can be observed from any given dataset. Therefore, let's assume we observe four unique data exemplars (worlds) denoted $\tau_1=\{A=1, B=1, C=1\}$, $\tau_2=\{A=2, B=2, C=2\}$, $\tau_3=\{A=3, B=2, C=1\}$, and  $\tau_4=\{A=1, B=3, C=3\}$ and are illustrated by the colored boxes in Figure ~\ref{fig:bkb-generalize}. Next, from these four worlds we learn four corresponding BKFs depicted by the colored arrows, where each colored arrow represents a S-node. By fusing these four fragments, our fused BKB allows us to generalize to unseen examples (the bottom illustration of Figure ~\ref{fig:bkb-generalize}) by combining S-nodes from different fragments to form a complete inference, weighted by each sources' reliability which, in the case our BKBSL implementation, are all the same. Therefore, using BKB fusion we can generalize to at least 3 more worlds that we have never been observed and cannot generalize to the world $\{A=1, B=2, C=1\}$ as there is no way to combine S-nodes from these four BKFs that yield a valid complete inference. 

\begin{algorithm}[t]
\caption{BKB-Fusion}\label{alg:bkb-fusion} 
\textbf{Inputs:} BKFs $K_1, K_2, \dots, K_n$ and Source Reliabilities $R$
\begin{algorithmic}[1]
\STATE Let $G'= (I', S', E')$ be an empty correlation graph
\STATE $I' \leftarrow \cup_i=1^nI_i$
\STATE $S' \leftarrow \cup_i=1^nS_i$
\STATE $E' \leftarrow \cup_i=1^nE_i$
\FOR{$K_i \in \{K_1, K_2, \dots, K_n\}$}
    \FOR{all S-nodes $q \in S_i$}
        \STATE Let $\alpha \leftarrow Head(q)$
        \STATE Let the source I-node for $q$ be $s = (S_{R_\alpha} = \sigma_i)$
        \STATE Add $s$ to $I'$ and add new S-node $q_s$ to $S'$
        \STATE Add edges $q_s \rightarrow s$ and $s\rightarrow q$ to $E'$
        \STATE Let $w'(q) \leftarrow w(q)$
    \ENDFOR
\ENDFOR
\FOR{all source variables $S_{R_\alpha}$}
    \STATE Let $\Lambda = \{s \mid s \text{ is a source node which is a state of } S_{R_\alpha}\}$ 
    \STATE $\rho \leftarrow \sum_{s\in\Lambda}r(s)$
    \FOR{$s\in \Lambda$, let $q_s$ be the S-node such that $q_s \rightarrow s \in E'$}
        \STATE Let $w'(q) \leftarrow \frac{r(s)}{\rho}$
    \ENDFOR
\ENDFOR
\STATE \textbf{return} $K' = (G', w')$
\end{algorithmic}
\end{algorithm}

\subsection{Additional Theorems}
\begin{theorem} \label{thm:max-mutual-info}
$C(P,Q)$ is a monotonically decreasing function of 
\begin{equation} \label{eq:sum-mutual-info}
    \sum_{i=1, \Pi(X_i)\neq \emptyset}^n W(X_i, \Pi(X_i))
\end{equation}
where 
\begin{equation}
    W(X_i, \Pi(X_i)) = \sum_{j=1}^{r_i}\sum_{k=1}^{c_i}p(x_{ij}, \pi_{ijk})\log_2\frac{p(x_{ij}, \pi_{ijk})}{p(x_{ij})p(\pi_{ijk})}
\end{equation}
Therefore, $C(P,Q)$ will be minimized if and only if the sum is maximized.
\end{theorem}

%% file: Tables/keel-mdl.tex
\begin{table}[ht]
\centering
\caption{\textit{Data} MDL results for BKBs learned with our BKB learning algorithm with GOBNILP DAG learning backend as well as BNs learned over same dataset with standard MDL scoring and GOBNILP. Also reported are the number of unique calls to the joint probability calculator for both BKB and BN learning.}
\begin{tabular}{lccccc}
\hline
Dataset           & BKB MDL          & BN MDL   & BKB Calls        & BN Calls & Calls Difference \\ \hline
newthyroid        & \textbf{-73}     & -462     & \textbf{1257}    & 7200     & -5943            \\
car               & \textbf{-994}    & -5976    & \textbf{20369}   & 40000    & -19631           \\
banana            & \textbf{-2047}   & -5929    & \textbf{103}     & 135      & -32              \\
poker             & \textbf{-976563} & -5263517 & \textbf{945}     & 946      & -1               \\
heart             & \textbf{-249}    & -1572    & \textbf{1666695} & 26262720 & -24596025        \\
saheart           & \textbf{-353}    & -1866    & \textbf{48005}   & 104976   & -56971           \\
wine              & \textbf{-134}    & -1103    & \textbf{1018988} & 39249360 & -38230372        \\
mammographic      & \textbf{-300}    & -2126    & \textbf{2735}    & 6750     & -4015            \\
pima              & \textbf{-628}    & -3059    & \textbf{28309}   & 43740    & -15431           \\
housevotes        & \textbf{-494}    & -1866    & \textbf{3209612} & 9746883  & -6537271         \\
tic-tac-toe       & \textbf{-1007}   & -4888    & \textbf{250986}  & 786432   & -535446          \\
winequality-red   & \textbf{-1087}   & -7423    & \textbf{663679}  & 4815936  & -4152257         \\
ecoli             & \textbf{-199}    & -875     & \textbf{9044}    & 108864   & -99820           \\
contraceptive     & \textbf{-1060}   & -6083    & \textbf{194374}  & 1080000  & -885626          \\
yeast             & \textbf{-391}    & -4624    & \textbf{47816}   & 380160   & -332344          \\
nursery           & \textbf{-7223}   & -59544   & \textbf{307592}  & 691200   & -383608          \\
titanic           & \textbf{-957}    & -2586    & \textbf{102}     & 108      & -6               \\
hayes-roth        & \textbf{-87}     & -403     & \textbf{928}     & 2000     & -1072            \\
iris              & \textbf{-69}     & -394     & \textbf{322}     & 1024     & -702             \\
bupa              & \textbf{-334}    & -760     & \textbf{1248}    & 2187     & -939             \\
haberman          & \textbf{-104}    & -384     & \textbf{80}      & 81       & -1               \\
winequality-white & \textbf{-2294}   & -24162   & \textbf{268095}  & 624312   & -356217          \\
glass             & \textbf{-155}    & -841     & \textbf{42539}   & 680400   & -637861          \\
flare             & \textbf{-394}    & -3407    & \textbf{275011}  & 9933408  & -9658397         \\
phoneme           & \textbf{-1265}   & -13925   & \textbf{12389}   & 55440    & -43051           \\
bands             & \textbf{-290}    & -1575    & \textbf{1038847} & 3064209  & -2025362         \\
australian        & \textbf{-507}    & -3793    & \textbf{1304495} & 6118368  & -4813873         \\
tae               & \textbf{-65}     & -263     & \textbf{540}     & 972      & -432             \\
post-operative    & \textbf{-61}     & -319     & \textbf{16202}   & 184320   & -168118          \\
vowel             & \textbf{-387}    & -4745    & \textbf{958224}  & 13288429 & -12330205        \\
balance           & \textbf{-505}    & -1503    & \textbf{312}     & 324      & -12              \\
zoo               & \textbf{-59}     & -650     & \textbf{1646540} & 37950312 & -36303772        \\
hepatitis         & \textbf{-82}     & -529     & \textbf{3634610} & 12986769 & -9352159         \\
crx               & \textbf{-242}    & -3788    & \textbf{1252575} & 7047840  & -5795265         \\
cleveland         & \textbf{-333}    & -1761    & \textbf{1536035} & 8398578  & -6862543         \\
lymphography      & \textbf{-109}    & -1083    & \textbf{2586093} & 33062432 & -30476339        \\
breast            & \textbf{-122}    & -1154    & \textbf{116161}  & 5225472  & -5109311         \\
led7digit         & \textbf{-122}    & -1599    & \textbf{10204}   & 24057    & -13853           \\
kr-vs-k           & \textbf{-7629}   & -78926   & \textbf{80933}   & 146875   & -65942           \\
monk-2            & \textbf{-229}    & -1548    & \textbf{6696}    & 8640     & -1944            \\ \hline
\end{tabular}
\label{tab:mdl-results}
\end{table}

%% file: Tables/keel-accuracy.tex
\begin{table}[ht]
\centering
\caption{Accuracy results on subset of benchmark datasets for both our learned BKBs and BNs using a 10-fold cross validation reported in terms of precision, recall and F1-score.}
\begin{tabular}{lccccccc}
\hline
               & \multicolumn{1}{l}{Precision} & \multicolumn{1}{l}{} & \multicolumn{1}{l}{Recall} & \multicolumn{1}{l}{} & \multicolumn{1}{l}{F1-Score} & \multicolumn{1}{l}{} & \multicolumn{1}{l}{\# Failed} \\ \cline{2-8} 
Dataset        & BKB                           & BN                   & BKB                        & BN                   & BKB                          & BN                   & BKB                           \\ \hline
newthyroid     & \textbf{0.900}                & 0.687                & \textbf{0.892}             & 0.809                & \textbf{0.886}               & 0.738                & 2                             \\
hayes-roth     & 0.572                         & \textbf{0.839}       & 0.481                      & \textbf{0.838}       & 0.448                        & \textbf{0.837}       & 0                             \\
iris           & \textbf{0.948}                & 0.940                & \textbf{0.947}             & 0.940                & \textbf{0.947}               & 0.940                & 0                             \\
tae            & \textbf{0.262}                & 0.143                & \textbf{0.397}             & 0.212                & \textbf{0.315}               & 0.171                & 0                             \\
haberman       & \textbf{0.744}                & 0.739                & 0.742                      & \textbf{0.758}       & 0.644                        & \textbf{0.743}       & 0                             \\
banana         & 0.704                         & \textbf{0.734}       & 0.698                      & \textbf{0.730}       & 0.688                        & \textbf{0.725}       & 0                             \\
titanic        & 0.458                         & \textbf{0.770}       & 0.677                      & \textbf{0.776}       & 0.547                        & \textbf{0.762}       & 0                             \\
led7digit      & \textbf{0.608}                & 0.016                & \textbf{0.644}             & 0.070                & \textbf{0.616}               & 0.026                & 0                             \\
monk-2         & \textbf{0.974}                & 0.870                & \textbf{0.972}             & 0.861                & \textbf{0.972}               & 0.859                & 0                             \\
balance        & \textbf{0.669}                & 0.522                & \textbf{0.726}             & 0.566                & \textbf{0.697}               & 0.543                & 0                             \\
mammographic   & 0.838                         & \textbf{0.841}       & 0.831                      & \textbf{0.841}       & 0.830                        & \textbf{0.841}       & 0                             \\
bupa           & 0.550                         & \textbf{0.630}       & 0.516                      & \textbf{0.629}       & 0.513                        & \textbf{0.629}       & 0                             \\
phoneme        & 0.767                         & \textbf{0.809}       & \textbf{0.779}             & 0.772                & 0.766                        & \textbf{0.781}       & 1                             \\
car            & \textbf{0.752}                & 0.490                & 0.277                      & \textbf{0.700}       & 0.187                        & \textbf{0.577}       & 0                             \\
post-operative & \textbf{0.586}                & 0.508                & 0.476                      & \textbf{0.713}       & 0.501                        & \textbf{0.593}       & 3                             \\
ecoli          & \textbf{0.595}                & 0.181                & \textbf{0.725}             & 0.426                & \textbf{0.651}               & 0.254                & 1                             \\
yeast          & \textbf{0.213}                & 0.097                & \textbf{0.354}             & 0.312                & \textbf{0.265}               & 0.148                & 0                             \\
pima           & \textbf{0.757}                & 0.689                & 0.672                      & \textbf{0.701}       & 0.677                        & \textbf{0.691}       & 0                             \\
glass          & \textbf{0.215}                & 0.126                & \textbf{0.430}             & 0.355                & \textbf{0.278}               & 0.186                & 0                             \\
saheart        & \textbf{0.730}                & 0.669                & 0.416                      & \textbf{0.684}       & 0.321                        & \textbf{0.670}       & 0                             \\
tic-tac-toe    & 0.589                         & \textbf{0.683}       & 0.652                      & \textbf{0.697}       & 0.554                        & \textbf{0.683}       & 9                             \\
contraceptive  & 0.051                         & \textbf{0.331}       & 0.226                      & \textbf{0.417}       & 0.083                        & \textbf{0.281}       & 0                             \\ \hline
\end{tabular}
\label{tab:accuracy}
\end{table}

%% file: Tables/keel-info.tex
\begin{table}[ht]
\centering
\caption{Benchmark dataset information including the maximum parent set limit used in our experiments for both BKB and BN learning.}
\begin{tabular}{lcccc}
\hline
Dataset           & Parent Set Limit & \# Features & \# I-nodes & \# Examples \\ \hline
newthyroid        & 5                & 6           & 21         & 215         \\
car               & 6                & 7           & 25         & 1728        \\
banana            & 2                & 3           & 14         & 5300        \\
poker             & 1                & 11          & 45         & 1025009     \\
heart             & 10               & 14          & 35         & 270         \\
saheart           & 9                & 10          & 22         & 462         \\
wine              & 8                & 14          & 40         & 178         \\
mammographic      & 5                & 6           & 21         & 830         \\
pima              & 8                & 9           & 21         & 768         \\
housevotes        & 7                & 17          & 34         & 232         \\
tic-tac-toe       & 9                & 10          & 29         & 958         \\
winequality-red   & 10               & 12          & 33         & 1599        \\
ecoli             & 7                & 8           & 29         & 336         \\
contraceptive     & 9                & 10          & 31         & 1473        \\
yeast             & 8                & 9           & 32         & 1484        \\
nursery           & 8                & 9           & 32         & 12960       \\
titanic           & 3                & 4           & 9          & 2201        \\
hayes-roth        & 4                & 5           & 18         & 160         \\
iris              & 4                & 5           & 15         & 150         \\
bupa              & 6                & 7           & 14         & 345         \\
haberman          & 3                & 4           & 8          & 306         \\
winequality-white & 4                & 12          & 45         & 4898        \\
glass             & 9                & 10          & 30         & 214         \\
flare             & 6                & 12          & 47         & 1066        \\
phoneme           & 5                & 6           & 34         & 5404        \\
bands             & 5                & 20          & 40         & 365         \\
australian        & 6                & 15          & 40         & 690         \\
tae               & 5                & 6           & 13         & 151         \\
post-operative    & 8                & 9           & 26         & 87          \\
vowel             & 5                & 14          & 60         & 990         \\
balance           & 4                & 5           & 11         & 625         \\
zoo               & 7                & 17          & 43         & 101         \\
hepatitis         & 6                & 20          & 40         & 80          \\
crx               & 5                & 16          & 54         & 653         \\
cleveland         & 10               & 14          & 31         & 297         \\
lymphography      & 5                & 19          & 63         & 148         \\
breast            & 9                & 10          & 43         & 277         \\
led7digit         & 7                & 8           & 24         & 500         \\
kr-vs-k           & 3                & 7           & 58         & 28056       \\
monk-2            & 6                & 7           & 19         & 432         \\ \hline
\end{tabular}
\label{tab:keel-info}
\end{table}